\documentclass[letterpaper, 10 pt, conference]{ieeeconf}  % Comment this line out if you need a4paper

\IEEEoverridecommandlockouts                              % This command is only needed if 
\usepackage{graphicx} % for pdf, bitmapped graphics files
\usepackage{amsmath} % assumes amsmath package installed
\usepackage{amssymb}  % assumes amsmath package installed
\usepackage[linesnumbered,ruled,vlined]{algorithm2e}
\usepackage{booktabs}
\usepackage{cancel}
\usepackage{subcaption}

\newtheorem{theorem}{Theorem}
\newtheorem{lemma}{Lemma}
\newtheorem{remark}{Remark}

\newtheorem{definition}{Definition}

\title{\LARGE \bf
On the similarities between Control Barrier Functions (CBFs) \\and Behavior Control Lyapunov Functions (BCLFs)
}

\author{Petter \"Ogren}% <-this % stops a space

\begin{document}

\maketitle
\thispagestyle{empty}
\pagestyle{empty}

%%%%%%%%%%%%%%%%%%%%%%%%%%%%%%%%%%%%%%%%%%%%%%%%
\begin{abstract}
Control Barrier Functions (CBFs) is an important tool used to address situations with multiple concurrent control objectives, such as safety and goal convergence.
In this paper we investigate the similarities between CBFs and so-called Behavior Control Lyapunov Functions (BCLFs) that have been proposed to address the same type of problems in the aeronautics domain.

The key results of both CBFs and BCLFs is the description of the set of controls that render a given set invariant.
We compare the corresponding theorems, and show that if we restrict the general class K function in CBFs to be the general linear function of BCLFs, and restrict the number of objectives as well as the number of priority levels to be just one in BCLFs, the results in terms of admissible control sets are equivalent. Furthermore, both papers show that the invariant set is  made asymptotically stable.

\end{abstract}

%%%%%%%%%%%%%%%%%%%%%%%%%%%%%%%%%%%%%%%%%%%%%%%%
\section{Introduction}

Many important control problems can be formulated in terms of two concurrent objectives: the need to reach a given \emph{goal} point while staying inside some \emph{safe} part of the statespace. This applies to autonomous cars, drones and other unmanned vehicles, but also to e.g. a manipulator arm grasping an object while staying away from joint limits as well as workspace obstacles.

Lyapunov theory \cite{sastry2013nonlinear} is a well known tool used to address the first of these objectives, goal convergence (including stability). However, it turns out that the same ideas of Lyapunov can also be used to address the second objective, staying within a safe set. This is the key observation behind two approaches that were developed independently, Behavior Control Lyapunov Functions (BCLFs) \cite{ogren2006autonomous} and Control Barrier Functions (CBFs)  \cite{ames2019control}.

BCLFs \cite{ogren2006autonomous}, were inspired by the ideas of handling concurrent objectives in  the subsumption architecture \cite{brooks1986robust} and the behavior based approach \cite{arkin1998behavior}, but strived to do so in a more formal way, with guarantees on performance.
This was done by applying the ideas of Control Lyapunov Functions (CLFs) \cite{sastry2013nonlinear}, but having several such functions in parallel, and placing bounds on the their time derivatives that depended on the available margins.

 Allowing for several concurrent objectives, BCLFs also include mechanisms for different priority levels, and acceptable safety margins for each of these levels and objectives. However, picking just one additional objective, and one priority level, the key result of BCLFs are very similar to those of CBFs, as will be shown in this paper.

CBFs were introduced in \cite{ames2014control}, and have since been applied to an ever increasing number of applications \cite{romdlony2014uniting,pickem2017robotarium,ames2016control,xiao2019decentralized,choi2020reinforcement,lindemann2020control}.

The main contribution of this paper is that we show that there are significant similarities between the theory of CBFs and BCLFs.
These similarities have not been shown before. In fact, the control community has been largely unaware of  BCLFs, as can be seen from the detailed description of the history of CBFs included in  \cite{ames2019control}. There it is stated that the first control version of a barrier certificate was presented in 2007, \cite{wieland2007constructive}, and that the  notion of extending $\dot h \geq 0$ at the boundary of the safe set to $\dot h \geq -h$ for the entire set was first done in 2009, \cite{aubin2009viability}. As shown below, both of these features were described in 2006, in \cite{ogren2006autonomous}, applying the constraint $\dot h \geq -k h$ for some $k>0$, which is similar in spirit to the more general
$\dot h \geq -\alpha(h)$ that was introduced in \cite{ames2014control,ames2016control}.

The outline of this paper is as follows. First we provide a background description of the key results of CBFs and BCLFs in Section II. Then we show the similarity of the two results in Section III, followed by conclusions in Section~IV.

%%%%%%%%%%%%%%%%%%%%%%%%%%%%%%%%%%%%%%%%%%%%%%%%

\section{Background}
In this section we will revise some key results on CBFs as well as BCLFs. For a complete description we refer the reader to  \cite{ames2019control} and  \cite{ogren2006autonomous}.

\subsection{Control Barrier Functions}
In \cite{ames2019control}, we have the following definitions and results.

\begin{definition}[From \cite{ames2019control}]
\label{cbf_1}
 Let $C \subset D \subset  \mathbb{R}^n$ be the superlevel set of a continuously differentiable function $h: d \rightarrow  \mathbb{R}$, then $h$ is a control barrier function (CBF) if there exists an extended class $K_\infty$ function $\alpha$ such that for $\dot x = f(x) + g(x)u$,
 \begin{equation}
 \sup_u\in U [L_f h(x) + L_g h(x) u] \geq -\alpha(h(x)) 
\end{equation}
 for all $x \in D$.
\end{definition}
Let the set of control values that render the set $C$ safe be defined as 
\begin{equation}
 K_{cbf}(x) = \{u \in U: L_f h(x) + L_g h(x) u + \alpha(h(x)) \geq 0\} \label{k_cbf}
\end{equation}

\begin{theorem}[From \cite{ames2019control}]
\label{th_cbf}
 Let $C  \subset \mathbb{R}^n$ be a set defined as the superlevel set of a continuously differentiable function  $h: D \subset \mathbb{R}^n \rightarrow \mathbb{R}$.
 If $h$ is a control barrier function on $D$ and $\frac{\partial h}{\partial x}(x) \neq 0$ for all $x \in \partial C$, then any Lipschits continuous controller $u(x) \in  K_{cbf}(x)$ renders the set $C$ safe. Additionally, the set C is asymptotically stable in $D$.
\end{theorem}

%%%%%%%%%%%%
\subsection{Behavior Control Lyapunov Functions}
 
 The ideas presented in \cite{ogren2006autonomous} is to define not a single safe set $\mathcal{C}=\{x: h(x) \geq 0\}$ as in \cite{ames2019control} but a family (over $j$ and $i$) of such sets $\{x: V_i(x) \leq b_{ij} \forall i\}$.
 Here $i$ indicates different types of objectives, such as obstacle clearance or time at target, whereas $j$ indicates different levels of ambition, such as having an obstacle clearance of 10 or 20 inches.

The initial definition of BCLFs is a bit different from Definition \ref{cbf_1}, since we are aiming for systems that can achieve objectives that are not met as well as keep those that are already met.
 
\begin{definition}[III.1 in \cite{ogren2006autonomous}]
\label{def_bclf}
 Given a system $\dot x = f(x,u)$, a bounded set of admissible controls $U_{adm}$ a piecewise $C^1$ function $V: \mathbb{R}^n \rightarrow  \mathbb{R} $,
 and scalars $b,\epsilon > 0$. Then $V$ is a BCLF for the bound $b$ and $\epsilon$  if 
 \begin{equation}
 \min_{u \in U_{adm} } \nabla V^T f(x,u) \leq - \epsilon, \forall x : V(x) \geq b.
\end{equation}
\end{definition}
 \vspace{2mm}
 In order to keep track of the family of objectives, over $i,j$ and be able to make informed tradeoffs between them, we need to define a priority level, as illustrated in Table~\ref{tab_cpl} and defined below.

\begin{table}
 \caption{Example of mission priorities captured by the CPL. Priority 0 is alway satisfied. First priority (CPL 1) is then to have an obstacle clearance of at least 10in. Second priority (CPL 2) is to arrive at the target within 60s from the start, while satisfying the 10in objective. Third priority is to keep the arrival time, but increase the obstacle margin, and adding some ``other objective". Note that all inequalities need to be $\leq$ but here the first row has been multiplied by $-1$ for readability.}
 \begin{center}
 \begin{tabular}{||l c | c c c c||} 
 \hline
 $b_{ij}$  & & CPL 0 & CPL 1 & CPL 2 & CPL 3 \\  [0.5ex] 
 \hline\hline
 Obstacle separation & $V_1 \geq$ & $-\infty$ & 10in & 10in & 20in \\ 
 \hline
 Time at target & $V_2 \leq$ & $\infty$ & $\infty$ & 60s & 60s\\
  \hline
 (other objective) & $V_3 \leq$ & $\infty$ & $\infty$ & $\infty$ & 100\\ %[1ex] 
 \hline
\end{tabular}
\end{center}
\label{tab_cpl}
\end{table}

  \begin{definition}(III.3 in \cite{ogren2006autonomous})
(CPL) Given a set of BCLFs, $V_i$, and bounds, $b_{ij}$, such that $b_{i0}= \infty$, for all $i$. Let the Current Priority Level (CPL) or $j_{CPL}(x)$, be given by
\begin{equation}
j_{CPL}(x)=\max\{j:V_i(x) \leq b_{ij}, \forall i\}, \label{eq_cpl}
\end{equation}
i.e., $j_{CPL}(x)$ is the rightmost column of $\{b_{ij}\}$ (see Table \ref{tab_cpl}) where all constraints are satisfied.
\end{definition}
Note that demanding $b_{i0}= \infty$, for all $i$ guarantees that the CPL is always defined. Next comes $U_{sat}$ the BCLF set corresponding to $K_{cbf}$ for CBFs.

\begin{definition}[III.4 in \cite{ogren2006autonomous}]
Fix $k > 0$ and let $\epsilon > 0$ be the same as in Definition \ref{def_bclf}.  Given a set
of BCLFs, $V_i$, with corresponding bounds, $b_{ij}$. Let
\begin{equation}
U_{sat}(x) = \{u:\dot V_i(x,u) \leq \frac{1}{k}(b_{ij}-V_i(x)), j = j_{CPL}(x), \forall i   \} \label{u_sat}
\end{equation}
the set of controls satisfying the bounds of the CPL. Let furthermore 
\begin{equation}
I_{next}(x) = \{i:V_i(x) \geq b_{i(j+1)}, j = j_{CPL}(x)   \}
\end{equation} 
the set of objectives to be focused on and  
\begin{equation}
U_{inc}(x) = \{u:\dot V_i(x,u) \leq - \epsilon, \forall i \in   I_{next}(x) \}
\end{equation} 
the set of controls aiming to increase the CPL.
\end{definition}
Note that $U_{sat}(x) $ is the set of controls preserving the CPL, $I_{next}(x) $ are the indices of the bounds that are not satisfied at the next CPL, and $U_{inc}(x)$ is that set of controls that move the state towards satisfy those bounds.

More formally, we characterize the sets that guarantee the satisfaction of the CPL and the future increase in priority level as follows.

\begin{lemma}(III.5 in \cite{ogren2006autonomous})
If a system starts at $x(t_0) = x_0$, and the chosen controls u satisfy 
\begin{equation}
 u \in U_{sat}(x),
\end{equation}

then $j_{CPL}(x_0) \leq j_{CPL}(x), \forall t > t_0$  i.e. the CPL will not decrease. 
If furthermore
 $$
 u \in U_{sat}(x) \cap U_{inc},
 $$
then 
$j_{CPL}(x_0) < j_{CPL}(x)$  
will be satisfied in finite time, i.e. the CPL will increase.

\end{lemma}
\begin{proof}
 See \cite{ogren2006autonomous}.
\end{proof}

Finally, the asymptotical stability part of Theorem \ref{th_cbf} has its counterpart below.

\begin{remark}(III.6 in \cite{ogren2006autonomous})
The constant $k$ governs how fast a satisfied bound $b_{ij}$ can be approached. If the worst possible $u \in U_{sat}$ is chosen, then we have equality in the constraints, 
$\dot V_i(x, u) = \frac{1}{k}(b_{ij}-V_i(x))$, and $V_i$ approaches $b_{ij}$
exponentially, with the time constant $k$.
\end{remark}

%%%%%%%%%%%%%%%%%%%%%%%%%%%%%%%%%%%%%%%%%%%%%%%%

\section{Similarity of Results}

The key similartiy between  \cite{ogren2006autonomous} and  \cite{ames2019control} is the set of controls making the set invariant, 
 $K_{cbf}(x)$ in Equation (\ref{k_cbf}) and $U_{sat}(x)$ in Equation (\ref{u_sat}).

\begin{lemma}
 The key sets of control choices that guarantee satisfaction of the objective, $ K_{cbf}$ and $ U_{sat}(x)$ are identical, if the class K function $\alpha: \mathbb{R} \rightarrow \mathbb{R}$ is chosen to be a generic linear function $\alpha(x) = x/k $ for some $k>0$, and the family of sets in (\ref{eq_cpl}) are reduced to a single one, that is $i=1,j=1$, with $b_{ij}=b_{11}=0$.
\end{lemma}
   
\begin{proof}
 Having $\dot x = f(x)+g(x)u$, $V_i(x)=-h(x)$,   the user defined bound  $b_{ij}=0$, and the class k function $\alpha(y)=\frac{1}{k}y$ for some user defined constant $k>0$ we have that
\begin{align}
 K_{cbf}(x) &= \{u \in U: L_f h(x) + L_g h(x) u + \alpha(h(x)) \geq 0\}  \nonumber \\
 &=\{u \in U: \dot h(x)  + \alpha(h(x)) \geq 0\} \nonumber \\
 &=\{u \in U: \dot h(x)   \geq -\frac{1}{k}(h(x))\} \nonumber \\
 &=\{u \in U: -\dot V_i(x)   \geq \frac{1}{k}(V_i(x))\} \nonumber \\
 &=\{u \in U: \dot V_i(x)   \leq \frac{-1}{k}(V_i(x))\} \nonumber \\
 &= U_{sat}(x) 
\end{align}
\end{proof}

With this observation made, Lemma III.5 of   \cite{ogren2006autonomous} states that $u \in U_{sat}(x)$ implies that the CPL will not decrease, which means that $V_i(x) \leq b_{ij}, \forall i$ for a given $j$, which translates to $h(x) \geq 0$.

Therefore, Lemma III.5 is very similar to Theorem 2, stating that $u \in K_{cbf}$ implies $h(x) \geq 0$.
Finally, the asymptotic stability of the safe set is also noted in Remark III.6.  
  
%%%%%%%%%%%%
\section{Conclusion}

In this paper we have shown the significant overlap between the CBF and the BCLF. This is important not only for understanding how ideas have developed, but also for future work on extending CBFs to domains where more than two objectives are concurrently considered.

 %We then explored some different applications in \cite{ogren2008improved,ogren2011model,wang2014dual}, with the most important detail being 
%the connection to Quadratic Programming in \cite{wang2014dual} below:
%\begin{itemize}
% \item
%  From page 4 of  \cite{wang2014dual} : ``
%Lemma 1. Problem 2 is equivalent to the following QP
%\begin{align*}
% \min_u & c^T u + u^T Q u \\
% \mbox{s.t.} & A_{ie} u \leq k_{ie}(b_{ie} - f_{ie}) - h_{ie} \\
% &A_{e} u = k_{e}(b_{e} - f_{e}) - h_{e} 
%\end{align*}
%
%%minu cTu+uTQu (12) s.t. Aieu ≤ kie(bie − fie) − hie (13) Aeu = ke(be − fe) − he (14)
%where $c = \frac{df_j}{dq}$, and each row of $A_{ie}, A_e, b_{ie}, b_e, f_{ie}, f_e, h_{ie}, h_e$
%contains the corresponding parts of $\frac{df_i}{dq}, b_i, f_i, \frac{df}{dt}$ respectively.
%
%Remark 2. Note that there are very efficient ways of solving such QPs, even in quite high dimensions. Note also that if Q = 0 we have a Linear Programming problem (LP) which is also easily solvable."
%\end{itemize}
%\vspace{-4mm}
%\section{}
%\subsection{}

%\clearpage refs are included in page limit for RAL
\bibliographystyle{ieeetr}
\bibliography{barrier}

\end{document}